\documentclass[journal]{IEEEtran}

\usepackage[lined, ruled, linesnumbered, commentsnumbered]{algorithm2e}
\usepackage{amsmath}
\usepackage{cite}
\usepackage{color}
\definecolor{chred}{rgb}{0.8,0,0}
\definecolor{chgray}{rgb}{0.5,0.5,0.5}
\usepackage{graphicx}
\usepackage{caption}
\usepackage{dblfloatfix}
\usepackage{subfigure}
\usepackage{eqlist}
\usepackage{txfonts}
\usepackage{url}
\usepackage{footmisc}
\usepackage{booktabs}

\usepackage{multirow}
\begin{document}
\title{A Mid-level Planning System for\\ Object Reorientation}

\author{Weiwei~Wan,~\IEEEmembership{Member,~IEEE,}
        Hisashi~Igawa,
        Kensuke~Harada,~\IEEEmembership{Member,~IEEE,}
        Zepei Wu,
        Hiromu~Onda,~\IEEEmembership{Member,~IEEE,}
        Kazuyuki~Nagata,~\IEEEmembership{Member,~IEEE,}
        Natsuki~Yamanobe,~\IEEEmembership{Member,~IEEE}% <-this % stops a space
\thanks{Weiwei Wan, Kensuke Harada, Zepei Wu, Kazuyuki Nagata, Hiromu Onda,
and Natsuki Yamanobe are with National Institute of Advanced
Industrial Science and Technology (AIST), Japan. Hisashi Igawa is with Hokkaido
Research Organization, Japan. Kensuke Harada is also affiliated with Osaka
University, Japan. Zepei Wu is also affiliated with Tsukuba University.
{\tt\small wan-weiwei@aist.go.jp}}}

\markboth{Journal of \LaTeX\ Class Files,~Vol.~x, No.~x, xxxx~2016}%
{Shell \MakeLowercase{\textit{et al.}}: Bare Demo of IEEEtran.cls for IEEE Journals}

\maketitle

\begin{abstract}
This paper presents a mid-level planning system for object reorientation. It
includes a grasp planner, a placement planner, and a regrasp sequence solver.
Given the initial and goal poses of an object, the mid-level planning system
finds a sequence of hand configurations that reorient the object from the initial to
the goal. This mid-level planning system is open to low-level motion planning
algorithm by providing two end-effector poses as the input. It is also open to
high-level symbolic planners by providing interface functions like placing an
object to a given position at a given rotation. The
planning system is demonstrated with several simulation examples and real-robot
executions using a Kawada Hiro robot and Robotiq 85 grippers.
\end{abstract}


\IEEEpeerreviewmaketitle

\section{Introduction}

\begin{figure*}[!htbp]
	\centering
	\includegraphics[width=\textwidth]{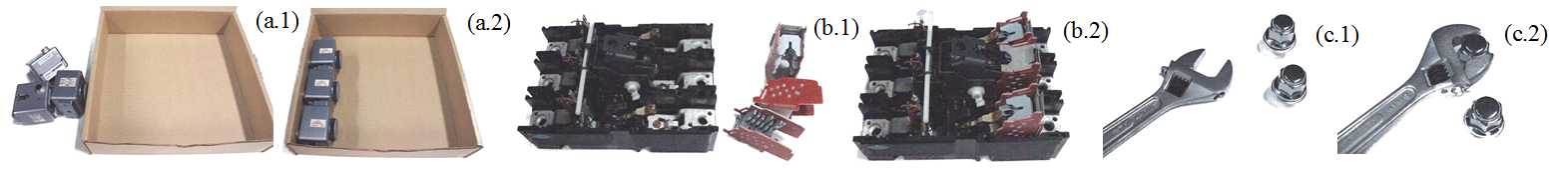}
	\caption{Some tasks that require object reorientation. (a) Packing
	objects. The task requires reorienting objects in (a.1) into the poses in (a.2). (b)
	Assembly. The task requires reorienting the capacitors in (b.1) into the
	poses in (b.2) and inserts them. (c) Using tools. The task requires
	reorienting the wrench in (c.1) into the pose in (c.2) so that
	its jaw fits into the nut.}
	\label{teaser}
\end{figure*}

\IEEEPARstart{O}{bject} reorientation is a common manipulation task for
robots. Given an initial pose of an object, a reorientation task
requires a robot to pick up the object, reorient it into a predefined pose,
and place it down at a certain position.

Some typical tasks that require object reorientation are shown
in Fig.\ref{teaser}.
The first one is a packing task. To finish this task, a robot is required to
reorient the object into the expected poses and pack them into the box. The
second one is an assembly task. A robot is required
to reorient the capacitor into the expected pose and insert it into the base.
The last one is to fastening a nut using a wrench. A robot must reorient the
wrench to have its jaw fit into the nut.
All these tasks are essentially object reorientation tasks.

The paper develops algorithms to
plan the motions that robots need to perform object reorientation. It presents a
mid-level planning system which includes a grasp planner, a
placement planner, and a regrasp sequence solver.
The input to the mid-level system is a sequence of goal poses of
the objects planned by a high-level assembly or symbolic planning component.
The output of the mid-level system is a sequence of robot poses and grasp
configurations that will be used by a low-level motion planning component.
Given the initial and goal poses of an object, the mid-level system finds a
sequence of robot poses and hand configurations that reorients the object from the
initial to the goal.
This mid-level system is open to low-level motion planning algorithms by
providing two end-effector poses as the start and goal. It is also open to
high-level symbolic or assembly planners by providing an interface function like
placing an object to a given position at a given rotation.

The essence of the mid-level planning system is a regrasp planner which
uses the stable placements of objects to increase the connectivity of grasp
configurations. To reorient the poses, a robot may place down the object at
intermediate placements and regrasp them. The regrasp reduces the constraints
from obstacles and robot kinematics.

The concept of mid-level planning system is novel in robotic planning community.
We in this paper develop a library that plays the role of the mid-level system
and demonstrate its efficacy with both simulation and real-world
experiments using a Kawada Hiro\footnote{http://nextage.kawada.jp/en/} robot and Robotiq 85 grippers.
The source code is available on GitHub\footnote{https://github.com/wanweiwei07/hiromatlab}.

The terminologies used are as follows. ``Object pose'' is used to refer to
the position and orientation of an object. ``Robot pose'' is used to refer to
a set of joint angles of the robot. ``Grasp configuration'' and ``regrasp
configuration'' are used to refer to a set of hand position, hand orientation
and hand joint angles.

\section{Background and Related Work}%

\begin{figure}[!htbp]
	\centering
	\includegraphics[width=.45\textwidth]{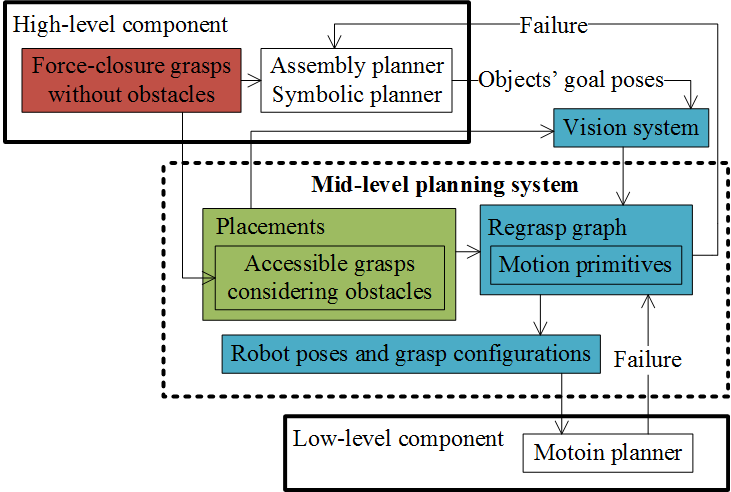}
	\caption{The position of the mid-level planning system is in the dashed
	framebox. The upper solid framebox shows the high-level component. The lower
	solid framebox shows the low-level component. The middle level system takes a
	sequence of goal poses of objects computed in the high-level
	component, computes a sequence of robot poses and grasp configurations, and sends them to the
	low-level motion component to find feasible motions.}
	\label{position}
\end{figure}

The position of a mid-level planning system in a planning platform is shown in
Fig.\ref{position}. In the highest level, the planning focuses on symbolic and
logical aspects. Examples include AND/OR graph assembly planning\cite{Mello:1990wqba}, geometric and physical reasoning
\cite{Mello91, Wilson94}, symbolic
deduction\cite{McDermott98, Hoffmann06}, etc. The high-level planners analyze
the contacts between objects, divide a task into sub-tasks, and decide the order
of operation. The output of the high-level component is usually a sequence of
goal poses of objects. In the low level, the planning focuses on the motion
between two configurations. Examples include probabilistic methods like Probabilistic
Roadmap (PRM)\cite{Kavraki96}, Rapidly-epxloring Random Tree
(RRT)\cite{Lavalle00}, etc., and optimization-based methods like Covariant
Hmiltonian Optimization for Motion Planning (CHOMP)\cite{Zucker12},
TrajOpt\cite{Schulman14}, etc. The low-level planners generate trajectories of
robots that avoid collision with obstacles and fulfill dynamic constraints. The
mid-level planning system takes the sequence of objects' goal poses
computed in the high-level component, computes a sequence of robot poses and grasp configurations, and
sends them to the low-level motion planner to find feasible motions. This paper
focuses on the mid-level aspect. It proposes the concept of mid-level planning
and develops libraries and real-world systems to demonstrate the concept.

To our best knowledge, the concept of mid-level planning is not strictly
defined, although related studies have been performed for decades. The most
relevant work is integrated task and motion planning.
The first integrated task and motion planning system is the STRIPS used in the
Shakey robot \cite{Fikes71}, which presented an integrated symbolic deduction
and movement planning system for a mobile robot. Following this seminal study,
much work has been devoted to the integrated task and motion planning for mobile
platforms. A good summary and discussion of the open problems before 2007 could
be found in \cite{Beta07}. More recent work like \cite{Lahijanian16} considers
temporal constraints and uncertainty in the environments. Integrated planners
are employed iteratively to generate safe motions for mobile robots to
finish office traversing tasks. \cite{Guo14} also does integrated task and motion
planning while considering temporal constraints and uncertainty. The task is
distributed to multiple robots. The sequence of executions as well as
motions are planned. \cite{KG09} only considers temporal constraints in the
integration and applies the integrated symbolic deduction and planning to plan the motions of mobile
robots in the presence of moving obstacles. \cite{Hertle11} considers
temporal constraints and applies the integrated planning to scenarios
like crew planning (multiple mobile robots) and room scanning.
These integrated planning in \cite{Lahijanian16, Guo14, KG09, Hertle11} are
about the tasks of mobile robots. The high-level planner finds motion sequences and the low-level planner finds
collision-free trajectories. Key poses of the mobile robots that connect the
high-level and low-level results are on a middle layer, which is
although implicit.

The planning systems on the tasks of manipulators also has an implicit middle
layer. \cite{Dogar15} presents an integrated assembly
and motion planning system which employs several mobile manipulators to assemble
a chair. The system has a high-level assembly sequence solver and a low-level
robot motion planner. The paper didn't explicitly claim a mid-level solver but was
actually using it to plan robot poses and grasp configurations. \cite{Bidot2015}
iteratively plans robot tasks and motions considering geometric constraints in
the environment. The system implicitly uses some mid-level planners to plan the
robot poses and grasp confiugrations considering preconditions.
\cite{Knepper:2013fn} presents an integrated planning system which plans
assembly sequences using geometric reasoning, plans coordinated manipulation
sequences using symbolic deduction, and plans robot poses and grasp
configurations considering each manipulation sequence \cite{Stein11} and robot
motions using sampling-based methods \cite{Knepper:2012}. The assembly sequences
and coordinated manipulation sequences are the result of high-level planners,
the robot motions are the results of low-level planners. The robot
poses and grasp configurations are on a middle layer which is between the high-level and low-level results.
\cite{Kaelbling13} presents an integrated symbolic deduction system which integrates viewpoint
planning, state estimation, and action planning. The advantage is that
uncertainty is considered when dealing with various geometric constraints.
Again, the robot poses and grasp configurations are in a middle level.
\cite{Dantam13} presents a similar integrated symbolic deduction and action
planning planning system. The concept of motion grammar is proposed as the
high-level interface. The touching poses are also the results of some
middle level planners. \cite{Kris10} presents an integrated task and PRM motion
planner which simultaneously samples in the sub-task space and transition space.
The transition space is actually what a mid-level planner should explore.
Many other studies like \cite{Dantam16, Bekris16, ICRA2014:Nedunuri,
King2013, Heger10, Eiichi10} are also implicitly employing some mid-level
planners.

% The role of a mid-level planning system is to facilitates the connections
% between the task planning in the high level and the motion planning
% in the low level. 
This paper explicitly develops a mid-level planning system for object
reorientation. Like the literature on integrated task and
motion planning, the sequences planned by our mid-level system are robot
poses and grasp configurations. The essence the mid-level planning system is a
regrasp planner which uses the stable placements of objects to increase the
connectivity of grasp configurations.
The seminal study of regrasp planning is \cite{Pierre87} which builds a
grasp-placement table (GP) and searches the table to find a sequence of grasping
and releasing key poses for object reorientation.
Some ensuing work that also uses search tables includes \cite{Rohrdanz97,
Sascha99, Hajime98, Cho03} which concentrates on industrial grippers and
\cite{Repela02, Zhixing08} which applies regrasp to dexterous hands. The most
up-to-date study based on the same idea is \cite{Lert16}.
Other work like \cite{Yoshihito92, Yoshihito94a, Niko09, Benjamin10, Harada14}
also do regrasp planning but the focus is more on planning of the sequences
rather than object reorientation. Our group published several articles on
regrasp \cite{Wan2015a, Wan2015c, Chao16}. These studies are based on graph
search rather than tables. Graph search has much higher performance and
makes it possible to plan dual-arm or multi-arm regrasps \cite{Jean10, Wan2016ral, Wan2016ar}.
This paper further develops our graph-based regrasp planner and uses it to
construct a mid-level planning system. The system has interfaces to the
high-level assembly or symbolic planners and low-level motion planners, and is demonstrated
with various object reorientation tasks like packing objects, assembly, and using
tools.

\section{System Overview}

\begin{figure}[!htbp]
	\centering
	\includegraphics[width=.45\textwidth]{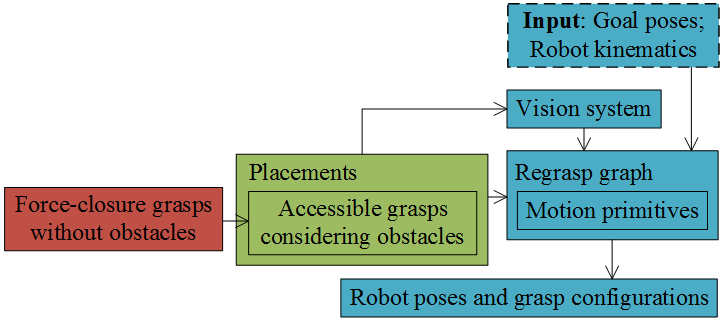}
	\caption{The red, green, and blue boxes indicate the grasp planner, the
	placements planner, and the regrasp sequence solver that compose the mid-level
	planning system. They correspond to the ones marked with the same color in
	Fig.\ref{position}. The input to the system includes the goal poses of
	the object and the kinematic parameters of the robots, it is in the
	upper-right corner and didn't appear in
	Fig.\ref{position}.}
	\label{midsys}
\end{figure}

Fig.\ref{midsys} shows an overview of the mid-level planning system. The colored
boxes in it correspond to the ones marked with the same colors in
Fig.\ref{position}. The mid-level planning system includes a grasp planner, a
placements planner, and a regrasp sequence solver which are marked with red,
green, and blue respectively. The input to the system includes the goal poses of
the object and the kinematic parameters of the robots, which are shown in the
upper-right corner of Fig.\ref{position}.

First, the system computes the force-closure grasps of an object using its geometric model
and the model of a given robotic gripper in the grasp planner.
Collision with obstacles are not considered in step. The output is a set of
poses and configurations of the gripper described in the object's local
coordinate system. Then, in the placement planner, the system finds the
placements of the object by computing its convex hull and checks on which facet
of the convex hull can the object stand stably. Each stable stand is one
placement and is described in the world coordinate system. The force-closure
grasps computed in the first step will be associated with the placements using
coordinate transformation between world coordinate system and the object's local
coordinate system, and collision detection with obstacles in the environment.
The output of the placement planner is the stable placements of the object
together with the accessible grasps to hold the placements. The regrasp sequence
solver uses the placements and their accessible grasps to find a sequence of
robot poses and grasp configurations that reorient the object from a given
start to a given goal. The initial pose of the object is assumed to be obtained
from a vision system. The goal poses of the object and the kinematic parameters
of the robot are input to the system by human beings.
The input are used together with the placements and their accessible grasps
to build a regrasp graph. The algorithms search the graph to compute a sequence
of robot poses and grasp configurations.

This mid-level planning system is
open to low-level motion planning algorithm by providing robot poses and grasp
configurations as the start and goal. The output of the system is a sequence of
robot poses and grasp configurations. Each adjacent pair of the poses and
configurations will be used as the start and goal of a motion planning algorithm
which provides direct interface to low level motion planners. On the other hand,
the system is open to high-level symbolic planners by providing interface
functions like placing an object to a given position at a given pose. The input
to the system is the initial and goal poses of the object.
The system solves regrasp sequences and decides the grasp and regrasp
configurations by itself in this mid-level. It blinds programmers from both
itself and the low-level motion planning details.

\section{Implementation Details}

This section introduces the implementation details of each part shown in
Fig.\ref{midsys}.

\subsection{Grasp planner}

We use the same grasp planning algorithm
presented in \cite{Wan2015a}. The robotic hand is supposed to be an industrial
gripper (Robotiq 85\footnote{http://robotiq.com/products/adaptive-robot-gripper/}). The process is
done without considering external collisions. Only the collision detection
between the hands and the object is performed to ensure feasibility. The output
of the grasp planner is a set $\textbf{G}=\{\textit{\textbf{g}}_1,
\textit{\textbf{g}}_2, \ldots, \textit{\textbf{g}}_n\}$ where an element
$\textit{\textbf{g}}_x=\{p_x, R_x, \textit{\textbf{j}}_x\}$. Here, the tuple
$(p_x, R_x)$ indicates the position and orientation of the robotic gripper.
The vector $\textit{\textbf{j}}_x$ indicates the angles of each finger joints.
Each $\textit{\textbf{g}}_x$ is computed following the requirements of force
closure using a predefined friction coefficient. Collision detection is
performed between $\textit{\textbf{g}}_x$ and the mesh model $\textbf{M}$ to
ensure a grasp doesn't collide the object.

\begin{figure}[!htbp]
	\centering
	\includegraphics[width=.45\textwidth]{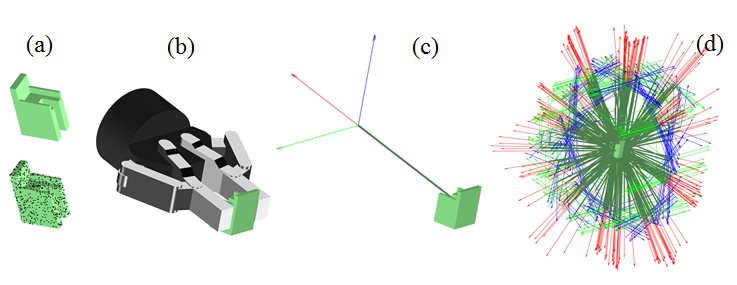}
	\caption{The grasp planner. The input of the grasp planner is an object
	model and a hand model shown. In (a), the object model is analyzed and sampled.
	In (b), the planner poses the hand at each pair of the samples on parallel
	facets and checks its force closure and collision with the object. The result
	is represented by a segment plus a coordinate system attached shown in (c).
	(d) shows all the grasps planned by the planner.}
	\label{grpplan}
\end{figure}

Fig.\ref{grpplan} shows the result of the grasp planner. Each grasp is
represented by a segment plus a coordinate system attached to the end of the
segment as shown in Fig.\ref{grpplan}(c) and (d).
The end of the segment is the $p_x$. The coordinate system implies the $R_x$.
The $\textit{\textbf{j}}_x$ is not illustrated.

\subsection{Placement planner}

Placement planner plans the stable placements of an object on a table surface
and associates the accessible grasps to the placements. We are using an improved
version of the placement planner presented in \cite{Wan2016ar}. In the previous
work, the placement planner computes the convex hull of the object and checks on
which facet of the convex hull can the object stand stably. Each stable stand is
treated as one placement of the object. Grasps are associated with a placement
considering collision with a horizontal supporting surface, which is the only
obstacles taken into account. In this improved version, we allow arbitrary obstacles.
Like the previous work, we compute the convex hull of the object's mesh model,
and find its stable placements on table surfaces. The difference is we allow
taking any rigid body obstacles into account. When associating the grasps, we
could not only specify a horizontal surface as an obstacle, but also arbitrary
rigid mesh models in the environment.

The output of the placement planner is a set
$\textbf{P}=\{\textit{\textbf{p}}_1, \textit{\textbf{p}}_2, \ldots,
\textit{\textbf{p}}_n\}$ where an element $\textit{\textbf{p}}_i=\{(p_i)^p,
(R_i) ^p, (\textbf{G}_i)^p\}$. Here, $((p_i) ^p, (R_i) ^p)$ indicates the
position and orientation of the object. $(\textbf{G}_i)^p$ indicates the
accessible force-closure and collision-free grasps that associate with the
object resting at pose $((p_i) ^p,( R_i) ^p)$. $(\textbf{G}_i)^p
=\{(\textit{\textbf{g}}_{i1})^p , (\textit{\textbf{g}}_{i2})^p , \ldots,
(\textit{\textbf{g}}_{in})^p\}$ where each $(\textit{\textbf{g}}_{ix})^p$ is
basically an element from the set $\textbf{G}$ computed by the grasp planner. It
is transformed to the coordinates of $\textit{\textbf{p}}_i$ using
$(\textit{\textbf{g}}_{ix})^p=(R_i)^p\cdot\textit{\textbf{g}}_x+(p_i)^p$.
Collision detection is performed between $(\textit{\textbf{g}}_{ix})^p$ and the
obstacles in the environment. Obstacles could be the supporting
table surfaces and some surrounding objects in the environment.

\begin{figure}[!htbp]
	\centering
	\includegraphics[width=.48\textwidth]{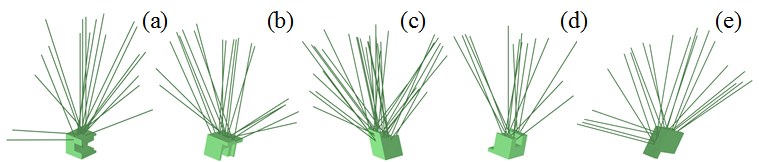}
	\caption{The placements of the object shown in Fig.\ref{grpplan}(a). Each
	placement is a stable state of the object standing on a horizontal surface and
	its accessible grasps. In this figure, there are no obstacles surrounding the
	placements and collision detection is only performed against the surface. The
	coordinates at the end of each segment representing the grasps are not shown.}
	\label{plcplan1}
\end{figure}

Fig.\ref{plcplan1} shows the placements of the object in Fig.\ref{grpplan}(a).
The object can stand on a table with six stable placements. The grasps
associated with the placements are shown using segments. The coordinates
attached to the end of segment are removed to ensure better visualization. There
are no obstacles surrounding the placements and collision detection is only
performed between $(\textit{\textbf{g}}_{ix})^p$ and the table surface. The
collided grasps are inaccessible and are not associated with the placements.

\begin{figure}[!htbp]
	\centering
	\includegraphics[width=.48\textwidth]{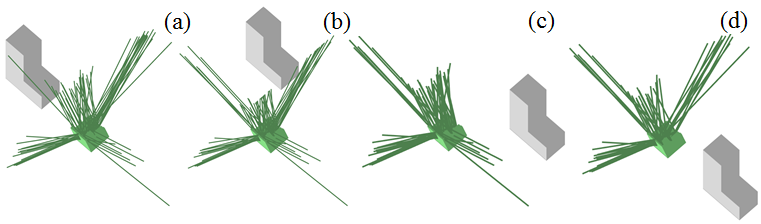}
	\caption{The changes of accessible grasps associated with a placement as
	surrounding obstacles change.}
	\label{plcplan2}
\end{figure}

Fig.\ref{plcplan2}(b) shows the cases where surrounding obstacles exist. The
gray block is supposed to be an obstacle. As the obstacle changes its position, the
accessible grasps associated with the object change correspondently. Collision
detection is only between both $(\textit{\textbf{g}}_{ix})^p$ and the table
surface, and both $(\textit{\textbf{g}}_{ix})^p$ and the obstacle.

\subsection{Regrasp sequence solver}

The regrasp sequence solver uses the placements and their associated grasps to
find a sequence of robot poses and grasp configurations that reorient the
object from a given start to a given goal. In implementation, the solver further examines the
accessible grasps considering the kinematics of a robot and motion
primitives\cite{Hauser06, Ding13}, builds a two-layer regrasp graph, connects
the start and goal to the graph, and searches the graph.

\subsubsection{Kinematics of the robot}

\begin{figure}[!htbp]
	\centering
	\includegraphics[width=.48\textwidth]{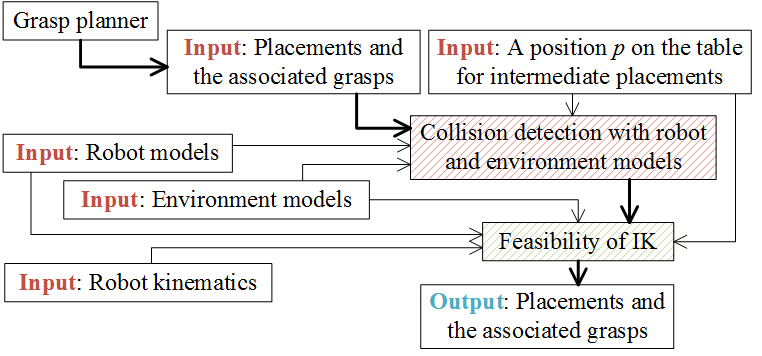}
	\caption{Further examine the grasps to ensure the accessibility of specific
	robots. In the red shadow box, the algorithm moves a placement to a specified
	position $p$. It examines the associated grasps and removes those grasps that
	collide with the environment and robot models. In the green box, the algorithm
	checks the feasibility of the robot's IK by posing its end effector at the
	remaining grasps. The output is the placements and their associated
	collision-free and IK-feasible grasps at $p$.}
	\label{ikplcplan}
\end{figure}

The grasps in previous sub-sections are computed without considering robot
kinematics. Before building the graph and searching for a regrasp sequence, the
grasps are further examined to ensure the accessibility to specific robots. The
flowchart in Fig.\ref{ikplcplan} shows how the examinations are performed.

Given a position $p$ on the table, the algorithm moves the placements to it and
checks the collision with surrounding obstacles. This collision detection step
is marked with red shadow in Fig.\ref{ikplcplan}. Collided grasps at the new $p$
are removed by this step. Then, for each of the remaining grasps, the
algorithm checks the feasibility of the robot's IK by posing its end effector at
them. The IK-infeasible grasps are also removed. The output of the algorithm
is still a set $\textbf{P}=\{\textit{\textbf{p}}_1, \textit{\textbf{p}}_2, \ldots,
\textit{\textbf{p}}_n\}$ where an element $\textit{\textbf{p}}_i=\{(p_i) ^p,
(R_i)^p, (\textbf{G}_i)^p\}$. Nevertheless, the $(p_i)^p$ is set to the newly
given $p$ and the elements of $(\textbf{G}_i)^p$ are both collision-free and IK-feasible.

\begin{figure}[!htbp]
	\centering
	\includegraphics[width=.48\textwidth]{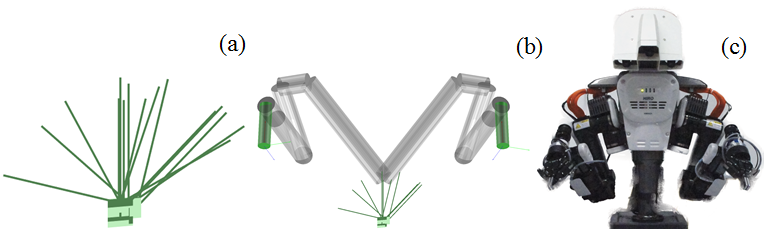}
	\caption{(a) A placement and its associated grasps. (b) The robotic kinematic
	model and IK-feasible grasps associated with the placement. (c) The Kawada
	Hiro robot.}
	\label{ikplcdemo}
\end{figure}

Fig.\ref{ikplcdemo} shows an example. A placement and its associated grasps
without considering robotic kinematics is shown in Fig.\ref{ikplcdemo}(a). In
contrast, the placement and its associated grasps at a position in front of a
Kawada Hiro robot, shown in Fig.\ref{ikplcdemo}(c)) is shown in
Fig.\ref{ikplcdemo}(b). The IK-infeasible grasps are removed.

\subsubsection{Incorporating motion primitives}

\begin{figure}[!htbp]
	\centering
	\includegraphics[width=.48\textwidth]{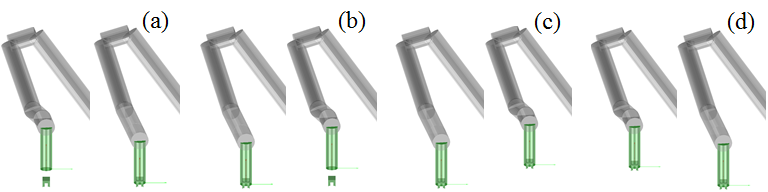}
	\caption{The four motion primitives: (a) The grasp
	primitive; (b) the release primitive; (c) the picking-up primitive; (d) the
	placing-down primitive.}
	\label{primitive}
\end{figure}

\begin{figure}[!htbp]
	\centering
	\includegraphics[width=.48\textwidth]{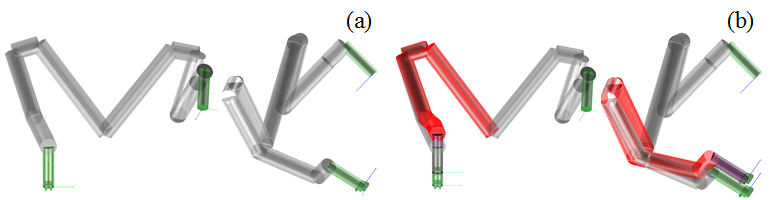}
	\caption{(a) Without considering motion primitives, the
	robot has to directly plan a motion between the initial and goal states, which
	involves contacts with the table and is a highly constrained
	problem. (b) Using motion planning, the robot only needs to plan between two
	intermediate states (red plots).}
	\label{planandprimitiveplan}
\end{figure}

Motion primitives are defined as two sequential IK-feasible grasps.
Fig.\ref{primitive} shows the four motion primitives used in the paper. It
includes: Fig.\ref{primitive}(a) A grasp primitive, which includes a pregrasp
key pose and a grasp key pose. The object pose doesn't change in the primitive;
Fig.\ref{primitive}(b) A release primitive, which includes a grasp key pose and
a pregrasp key pose. It is the inverse of (a) and the object pose doesn't change; Fig.\ref{primitive}(c) A
picking-up primitive, which includes a grasp key pose and a retraction key pose.
The object retracts together with the retraction key pose in the primitive;
Fig.\ref{primitive}(d) A placing-down primitive. It is the inverse of (c).
The reason to incorporate motion primitive is to relax motion planning. Without
considering the primitives, the robot has to plan a motion between the initial
and goal states shown in Fig.\ref{planandprimitiveplan}(a) which involves
contacts between the object and the table surface in workspace and narrow or
highly constrained configuration spaces in configuration space \cite{Hsu03,
Yershova2009}.
Motion primitives relaxes the planning around the
narrow configuration spaces. A robot only needs to plan between two intermediate
states. The motion between the intermediate states and the initial and goal
states are defined by the primitives. An example is shown in
Fig.\ref{planandprimitiveplan}(b), the picking-up and placing-down motion
primitives take care of the motion between the intermediate states (red plots)
and the initial and goal states, leaving motion planners to find a motion
between the intermediate states.

Incorporating motion primitives requires further examining their feasibility.
For a grasp $\textit{\textbf{g}}_{x} =(p_x, R_x, \textit{\textbf{j}}_x)$,  the
grasp and release motion primitives are defined as ($(p_x, R_x,
\textit{\textbf{j}}_x)$,  $(p_x+\omega[R_x(1,1), R_x(2,1), R_x(3,1)], R_x,
\textit{\textbf{j}}_x)$) and ($(p_x+\omega [R_x(1,1), R_x(2,1), R_x(3,1)], R_x,
\textit{\textbf{j}}_x)$, $(p_x, R_x, \textit{\textbf{j}}_x)$) where $[R_x(1,1),
R_x(2,1), R_x(3,1)]$ is the approaching direction of the robot gripper. $\omega$
controls the scale of the primitive. The picking-up and placing-down motion
primitives are defined as ($(p_x, R_x, \textit{\textbf{j}}_x)$, 
$(p_x+\omega\cdot backward, R_x, \textit{\textbf{j}}_x)$) and ($(p_x+\omega\cdot backward, R_x,
\textit{\textbf{j}}_x)$, $(p_x, R_x, \textit{\textbf{j}}_x)$) where $backward$ is
planned by a high-level planner. If $backward$ is the upward direction
$[0,0,1]$, the picking-up and placing-down motion
primitives are exactly along the up-down direction.
$\omega$ is the same controlling parameter.
The IK at $(p_x+\omega\cdot[R_x(1,1), R_x(2,1), R_x(3,1)], R_x,
\textit{\textbf{j}}_x)$ and $(p_x+\omega\cdot backward, R_x,
\textit{\textbf{j}}_x)$ are further examined to ensure the four motion
primitives are feasible. After incorporating motion primitives, each element of
the placement is saved as a triple
$\textit{\textbf{p}}_i=(\textit{\textbf{p}}_i^{o}, \textit{\textbf{p}}_i^{pre},
\textit{\textbf{p}}_i^{ret})$ where $\textit{\textbf{p}}_i^{o}$ is the original
placement and its associated grasps, $\textit{\textbf{p}}_i^{pre}$ is the
placement with the grasps for grasp and release motion primitives, and 
$\textit{\textbf{p}}_i^{ret}$ is the placement with the grasps for
picking-up and placing-down motion primitives.

\begin{figure}[!htbp]
	\centering
	\includegraphics[width=.48\textwidth]{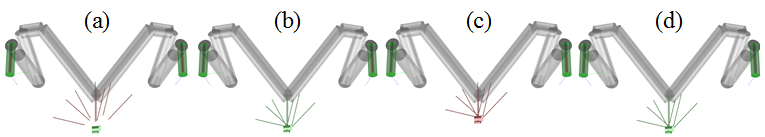}
	\caption{Accessible grasps after further considering motion primitives.
	Comparing with Fig.\ref{ikplcdemo} the number of accessible grasps are further decreased.
	(a) corresponds to $\textit{\textbf{p}}_i^{pre}$. (c) corresponds to
	$\textit{\textbf{p}}_i^{ret}$. (b) and (d) correspond to
	$\textit{\textbf{p}}_i^{o}$.}
	\label{gafterp}
\end{figure}

Fig.\ref{gafterp} shows the results after further reducing the IK-infeasible
grasps considering motion primitives. Comparing with Fig.\ref{ikplcdemo} the
number of accessible grasps are further decreased. Fig.\ref{gafterp}(b) and (d)
correspond to $\textit{\textbf{p}}_i^{o}$. Fig.\ref{gafterp}(a) corresponds to
$\textit{\textbf{p}}_i^{pre}$. Fig.\ref{gafterp}(c) corresponds to
$\textit{\textbf{p}}_i^{ret}$.

\subsubsection{The two-layer regrasp graph}

The regrasp graph connects the placements using the shared grasps. For two
placements $\textit{\textbf{p}}_i$ and $\textit{\textbf{p}}_j$ (more exactly
$\textit{\textbf{p}}_i^{o}$ and $\textit{\textbf{p}}_j^{o}$), if $\exists
\textit{\textbf{g}}_x$ where $(R_i)^p\cdot\textit{\textbf{g}}_x+(p_i)
^p\in\textbf{G}_i$ and $(R_j)^p\cdot\textit{\textbf{g}}_x+(p_j)
^p\in\textbf{G}_j$, they can be connected. It means there exists a shared grasp
that is identical in the object's local coordinate system and is associated with
both placements. A robot could pick up the object resting at
$\textit{\textbf{p}}_i$  using a grasp
$\textit{\textbf{g}}_{ix}=(R_i)^p\cdot\textit{\textbf{g}}_x+(p_i)^p$, transforms
its end effector pose to $(p_j, R_j)$, and places the object down with the grasp
$\textit{\textbf{g}}_{jx}=(R_j)^p\cdot\textit{\textbf{g}}_x+(p_j)^p$.

\begin{figure}[!htbp]
	\centering
	\includegraphics[width=.48\textwidth]{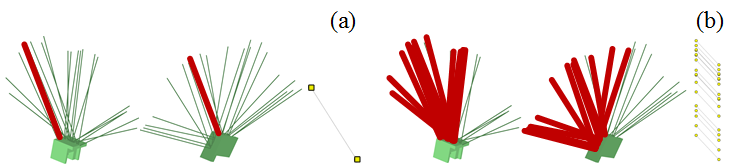}
	\caption{The shared grasp of two placements. (a) The first layer expresses
	connectivity. If there exists at least one common grasp, the two placements are
	connected with each other in the first layer. (b) The second layer expresses
	the details of the connection. Each shared grasp leads to one
	segments in the second layer.}
	\label{gtoggraph}
\end{figure}

Fig.\ref{gtoggraph}(a) shows this shared grasp using a red segment. These two
placements in the figure are represented as two rectangular nodes and are
connected with each other (see the right part of (a)). This connection is in the first layer
which only shows whether two placements are connectible. The two
rectangular nodes correspond to ($p_i$, $R_i$) and ($p_j$, $R_j$)
respectively. Grasps are not involved.
The second layer of the graph further shows the number of shared grasps. See
Fig.\ref{gtoggraph}(b) for examples. The two placements share many grasps (see
the red segments). Each shared grasp leads to one segment between the two
placements in the second layer. The two circular nodes at the end of the
segment correspond to $(R_i)^p\cdot\textit{\textbf{g}}_{x_k}+(p_i)
^p$ and $(R_j)^p\cdot\textit{\textbf{g}}_{x_k}+(p_j)^p\in\textbf{G}_j$
respectively. The subscript $k=1,2,\ldots,n$ indicates the number of shared
grasps.

\begin{figure}[!htbp]
	\centering
	\includegraphics[width=.41\textwidth]{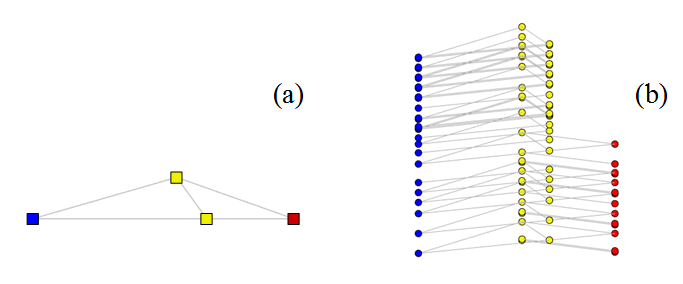}
	\caption{(a) The first layer of the regrasp graph, which shows the
	connectivity of initial pose (blue), goal pose (red), and the placements
	(yellow). (b) The second layer of the regrasp graph, which holds the details of
	each shared grasp.}
	\label{thegraph}
\end{figure}

All placements are connected to the graph like this. The initial and goal poses
of the object are later connected to the graph composed by placements for
searching. The graph of the exemplary object in Fig.\ref{grpplan} is show in
Fig.\ref{thegraph}. Here, the initial and goal poses are the two placements
shown in Fig.\ref{gtoggraph}.
They are marked using blue and red nodes in the graph. The yellow nodes show the
intermediate placements. The algorithm searches the graph to find a shorted path
to reorient the object from its initial pose to the goal pose. The path may
include the yellow nodes, which means the robot has to place down the object at
an intermediate placement to do regrasp. 

Note that all nodes on the graph, be it the
initial pose, goal pose, or the intermediate placements, are essentially
stable placements on the table. Consequently, the nodes in the graph correspond
to $\textit{\textbf{p}}_i^{o}$.
$\textit{\textbf{p}}_i^{pre}$ and $\textit{\textbf{p}}_i^{ret}$ are not
explicitly shown. A graph search algorithm searches the graph and find a path
composed by $\textit{\textbf{p}}_i^{o}$. $\textit{\textbf{p}}_i^{pre}$ and
$\textit{\textbf{p}}_i^{ret}$ are added to each node of the path afterwards to
include motion primitives. Alg.\ref{searchexp} shows this algorithm. It includes
two parts. The first part is graph searching. First, the algorithm searches the
first layer of the graph to find a sequence of $\textnormal{(}p_i, R_i\textnormal{)}$.
Then, for each adjacent pair of $\textnormal{(}p_i, R_i\textnormal{)}$ and
$\textnormal{(}p_{i+1}, R_{i+1}\textnormal{)}$, the algorithm finds a shared
grasp $\textit{\textbf{g}}_{x_i}$ of them in the second layer. The output of the
first part is a sequence $\textit{\textbf{p}}_{1}^{o}$,
$\textit{\textbf{p}}_{2}^{o}$, \ldots, $\textit{\textbf{p}}_{2n-2}^{o}$ where
two adjacent elements ($\textit{\textbf{p}}_{2i-1}^{o}$,
$\textit{\textbf{p}}_{2i}^{o}$, $i=1,2,\ldots,n-1$) are
$\textit{\textbf{p}}_{2i-1}^{o}$=$\textnormal{(}p_{i}, R_{i},
\textit{\textbf{g}}_{ix_i}\textnormal{)}$ and
$\textit{\textbf{p}}_{2i}^{o}$=$\textnormal{(}p_{i+1}, R_{i+1},
\textit{\textbf{g}}_{{i+1}x_i}\textnormal{)}$. The second part expands the
results of the first part by including $\textit{\textbf{p}}_i^{pre}$ and
$\textit{\textbf{p}}_i^{ret}$. The two elements in an adjacent pair
($\textit{\textbf{p}}_{i}^{o}$, $\textit{\textbf{p}}_{i+1}^{o}$), will be
expanded to ($\textit{\textbf{p}}_{i}^{pre}$, $\textit{\textbf{p}}_{i}^{o}$,
$\textit{\textbf{p}}_{i}^{ret}$, $\textit{\textbf{p}}_{i+1}^{ret}$, $\textit{\textbf{p}}_{i+1}^{o}$,
$\textit{\textbf{p}}_{i+1}^{pre}$). The robot picks up an object using a grasp
primitive $\textit{\textbf{p}}_{i}^{pre}$$\rightarrow$
$\textit{\textbf{p}}_{i}^{o}$ and a picking-up primitive
$\textit{\textbf{p}}_{i}^{o}$$\rightarrow$
$\textit{\textbf{p}}_{i}^{ret}$, and
places it down using a placing down primitive
$\textit{\textbf{p}}_{i+1}^{ret}$$\rightarrow$
$\textit{\textbf{p}}_{i+1}^{o}$
and a release primitive
$\textit{\textbf{p}}_{i+1}^{o}$$\rightarrow$
$\textit{\textbf{p}}_{i+1}^{pre}$.

\begin{algorithm}[!htbp]
  \SetKwData{Null}{null}
  \SetKwFunction{dijkstra}{dijkstra}
  \SetKwFunction{sharedGrasp}{sharedGrasp}
  \SetKwFunction{append}{append}
  \SetKwFunction{expandPrimitive}{expandPrimitive}
  \DontPrintSemicolon
  \KwData{The regrasp graph $\mathcal{G}$; $\mathcal{G}^1$ incidates the first
  layer; $\mathcal{G}^2$ indicates the second layer}
  \KwResult{A sequence of robot poses and grasp configurations}
  \Begin {
  	\textit{/*Part 1, search the graph*/}\\
  	$\textnormal{(}\textnormal{(}p_1, R_1\textnormal{)}$,
  	$\textnormal{(}p_2, R_2\textnormal{)}$, $\ldots$,
  	$\textnormal{(}p_n, R_n\textnormal{)}\textnormal{)}\leftarrow$\dijkstra{$\mathcal{G}^1$}\\
  	\For{$i\in\{1,2,\ldots,n-1\}$}{
  		$\textnormal{(}\textit{\textbf{g}}_{ix_i}$,
  		$\textit{\textbf{g}}_{{(i+1)}x_i}\textnormal{)}$\\
		~~~~~~~~~~$\leftarrow$\sharedGrasp{\textnormal{(}$p_i$,
  		$R_i$\textnormal{)}, \textnormal{(}$p_{i+1}$, $R_{i+1}$\textnormal{)},
  		$\mathcal{G}^2$}\\
  		$\textit{\textbf{p}}_{2i-1}^{o}\leftarrow\textnormal{(}p_{i}, R_{i}, \textit{\textbf{g}}_{ix_i}\textnormal{)}$\\
  		$\textit{\textbf{p}}_{2i}^{o}\leftarrow\textnormal{(}p_{i+1}, R_{i+1}, \textit{\textbf{g}}_{(i+1)x_i}\textnormal{)}$\\
  	}
  	\textit{/*Part 2, include the motion primitives*/}\\
  	$\textnormal{sequence}\leftarrow\emptyset$\\
  	\For{$i\in\{1,3,5,\ldots,2n-3\}$}{
  		$\textit{\textbf{p}}_{i}^{pre}$,
  		$\textit{\textbf{p}}_{i}^{o}$,
  		$\textit{\textbf{p}}_{i}^{ret}\leftarrow$
  		\expandPrimitive{$\textit{\textbf{p}}_{i}^{o}$}\\
  		$\textit{\textbf{p}}_{i+1}^{pre}$,
  		$\textit{\textbf{p}}_{i+1}^{o}$,
  		$\textit{\textbf{p}}_{i+1}^{ret}\leftarrow$
  		\expandPrimitive{$\textit{\textbf{p}}_{i+1}^{o}$}\\
  		$\textnormal{sequence}$.\append{
  		$\textit{\textbf{p}}_{i}^{pre}$,
  		$\textit{\textbf{p}}_{i}^{o}$,
  		$\textit{\textbf{p}}_{i}^{ret}$,
  		$\textit{\textbf{p}}_{i+1}^{pre}$,
  		$\textit{\textbf{p}}_{i+1}^{o}$,
  		$\textit{\textbf{p}}_{i+1}^{ret}$}\\
  	}
    \Return{$\textnormal{sequence}$}\;
  }
  \caption{Graph searching and expanding}
  \label{searchexp}
\end{algorithm}

\begin{figure*}[!hbtp]
	\centering
	\includegraphics[width=.98\textwidth]{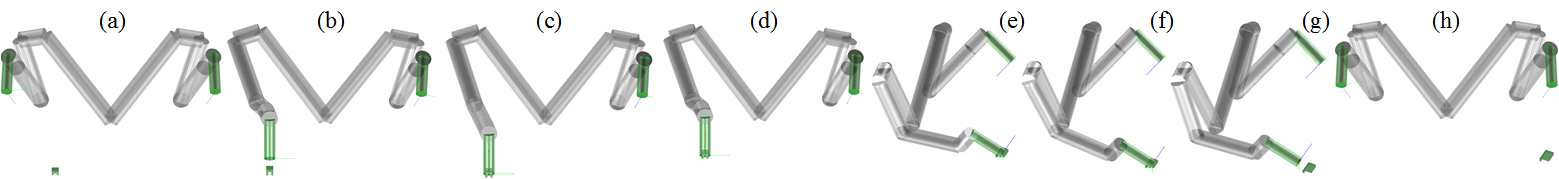}
	\caption{The results planned by the mid-level system. In this example, the robot can directly
	reorient the object from its initial pose to the goal, without employing
	intermediate placements and regrasps. The output
	of graph search in the first layer is (c)$\rightarrow$(f).
	The remaining (b), (c), \ldots, (g) are the expanded
	primitives. (b)$\rightarrow$(c) is the grasp primitive.
	(c)$\rightarrow$(d) is the picking-up primitive. (e)$\rightarrow$(f) is the
	placing down primitive. (f)$\rightarrow$(g) is the release primitive. (a) and
	(h) are the standard robot pose.}
	\label{finalseq}
\end{figure*}

Fig.\ref{finalseq} shows the results of the whole mid-level system. The output
of graph search in the first layer is (c)$\rightarrow$(f).  The robot can directly
reorient the object from its initial pose to the goal, without employing
intermediate placements and regrasps. (b), (d), (e), and (g) are the expanded
states considering motion primitives. (b)$\rightarrow$(c) is the grasp primitive. (c)$\rightarrow$(d)
is the picking-up primitive. (e)$\rightarrow$(f) is the placing down
primitive. (f)$\rightarrow$(g) is the release primitive. (a) and (h) are the
standard robot pose. The results can be used by motion planning in a lower
level to plan motions between (a) and (b), (d) and (e), and (g) and (h).

\section{Experiments and Analysis}

We demonstrate the efficacy of the planner using three
tasks: (1) Packing objects, (2) assembly, and (3) using tools. 
Both simulation and real-world experiments are performed.
The computer used in the simulation is a Thinkpad P70 mobile workstation with an
Intel(R) Xeon(R) CPU E3-1505M v5 \@ 2.80Hz and 32.0GB RAM. The robot used in the
real-world experiments is a Kawada Nextage Open. The objects and tools used in
the experiments are from ``Kenjin Puzzle'' and ``Battat Take-A-Part Toy Vehicles
Airplane''. They are available on Amazon\footnote{http://amzn.to/29YKoKX and http://amzn.to/2avUXFZ}.
A video of the experimental results is available in the supplementary material.

\subsection{Packing objects}

\begin{figure}[!htbp]
	\centering
	\includegraphics[width=.4\textwidth]{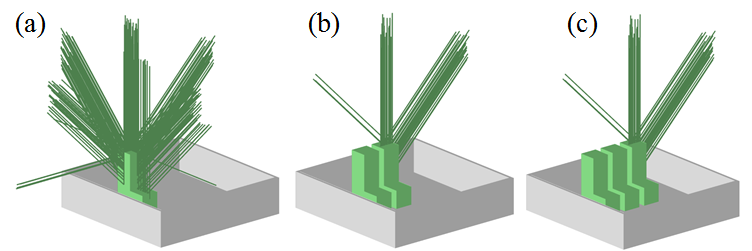}
	\caption{The accessible grasps of each object in the packing task. The packing
	sequence (a) $\rightarrow$ (b) $\rightarrow$ (c) is computed by a high level
	planning component. The grasp planner of the mid-level planning system finds
	the accessible grasps when packing objects following this sequence.}
	\label{graspexpsim1}
\end{figure}

The mesh models used in the packing objects task are shown in
Fig.\ref{graspexpsim1}. The goal is to pack three objects into a box. The goal
poses of the objects are given. A high level planning component computes a
packing sequence of the goal poses which will be sent to the mid-level planning
system as input. The packing sequence indicates which object to pack first and
which objects to pack later. The mid-level planner uses the packing sequence to compute the
manipulation sequence of each object.

First, the grasp planner of the mid-level planning system computes the
accessible grasps of each object. Fig.\ref{graspexpsim1}(a), (b), and (c) show
the accessible grasp. The packed objects become the obstacles of the remaining
objects, which leads to fewer accessible grasps in Fig.\ref{graspexpsim1}(b)
and Fig.\ref{graspexpsim1}(c).

\begin{figure}[!htbp]
	\centering
	\includegraphics[width=.46\textwidth]{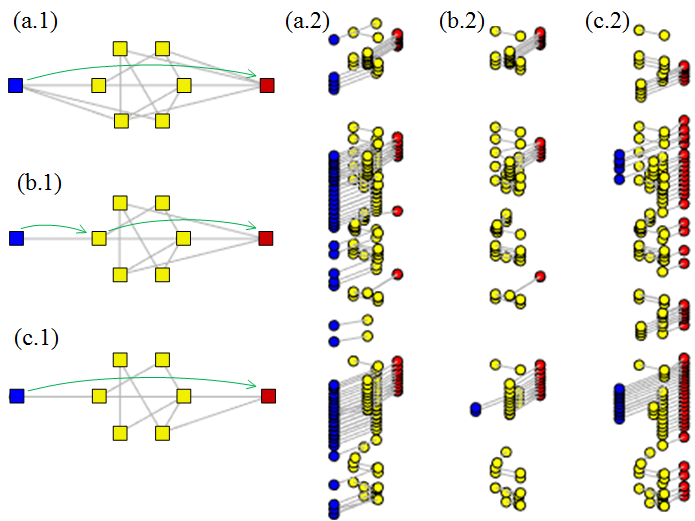}
	\caption{Regrasp graph of the three objects in the packing task. (a.1) and
	(a.2) are the first layer and second layer of the regrasp graph for the
	first object. (b.1) and (b.2) are the first layer and second layer of the
	regrasp graph for the second object. (c.1) and (c.2) are the first layer and
	second layer of the regrasp graph for the third object.}
	\label{packingobjgraphs}
\end{figure}

\begin{figure*}[!htbp]
	\centering
	\includegraphics[width=.98\textwidth]{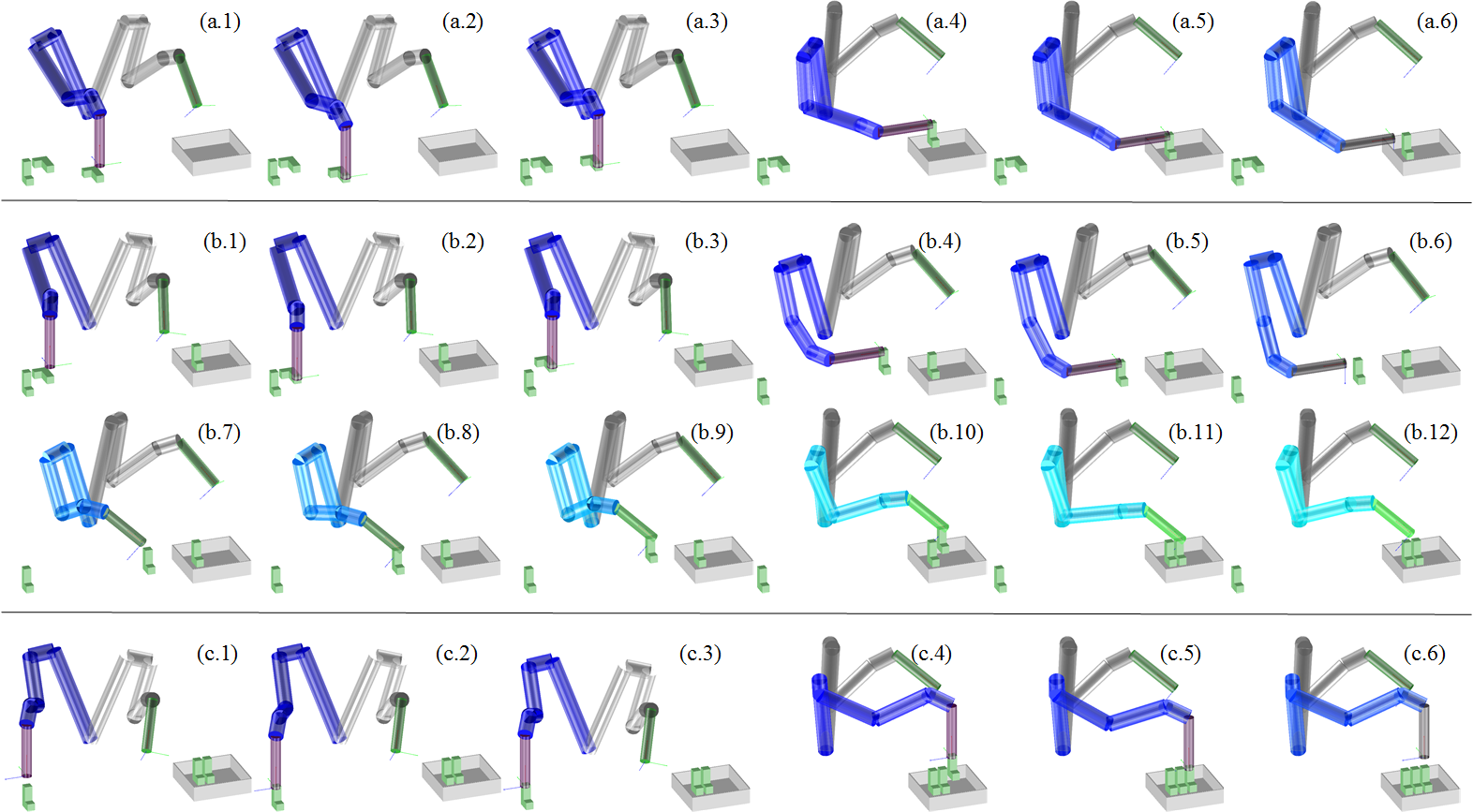}
	\caption{The simulation results of packing objects. (a.1), (a.2), \ldots,
	(a.3) are the robot poses and grasp configurations to pack the first object.
	(b.1), (b.2), \ldots, (b.12) are the robot poses and grasp configurations to
	pack the second object. (c.1), (c.2), \ldots,
	(c.3) are for packing the third object.}
	\label{packingobjects}
\end{figure*}

Then, the placement planner computes the stable placements for the object and
the regrasp solver uses the stable placements to build a regrasp graph. The
regraspgraphs are built for each object and are shown in
Fig.\ref{packingobjgraphs}. Fig.\ref{packingobjgraphs}(a.1) and
(a.2) are the first layer and second layer of the regrasp graph for the
first object. Fig.\ref{packingobjgraphs}(b.1) and (b.2) are the first layer and
second layer of the regrasp graph for the second object.
Fig.\ref{packingobjgraphs}(c.1) and (c.2) are the first layer and second layer of the regrasp graph for the third object.
The regrasp solver searches the regrasp graphs to find regrasp sequences. The
output path on the first layers are marked using green arrows in
Fig.\ref{packingobjgraphs}(a.1), (b.1), and (c.1), which implies packing the
first object and the third object can be done directly, whereas packing the
second object requires one time of regrasp.
The sequence of robot poses and grasp configurations for the three objects are
shown in Fig.\ref{packingobjects}. The initial and goal poses of the objects
are shown in Fig.\ref{packingobjects}(a.1) and (c.6). The sequence (a.1), (a,2), \ldots,
(a.6) is to pack the first object.
Fig.\ref{packingobjects}(a.2) $\rightarrow$ (a.5) corresponds to the green arrow
in Fig.\ref{packingobjgraphs}(a.1).
The remaining poses and configurations in Fig.\ref{packingobjects}(a.1), (a.3),
(a.4), and (a.6) are the extended primitives. The sequence
Fig.\ref{packingobjects}(b.1), (b,2), \ldots, (b.12) is to pack the second
object. Fig.\ref{packingobjects}(b.2) $\rightarrow$ (b.5)(b.8) $\rightarrow$
(b.11) corresponds to the green arrows in Fig.\ref{packingobjgraphs}(b.1). The
remaining poses and configurations
Fig.\ref{packingobjects}(b.1), (b.3), (b.4), (b.6), (b.7), (b.9), (b.10), and
(b.12) are the extended primitives.
The sequence Fig.\ref{packingobjects}(c.1), (c.2), \ldots, (c.6) is to pack the
third objects where Fig.\ref{packingobjects}(c.2) $\rightarrow$ (c.5)
corresponds to the green arrow in Fig.\ref{packingobjgraphs}(c.1). The remaining robot
poses and grasp configurations in Fig.\ref{packingobjects}(c.1), (c.3), (c.4),
and (c.6) are the extended primitives.

\begin{figure}[!htbp]
	\centering
	\includegraphics[width=.48\textwidth]{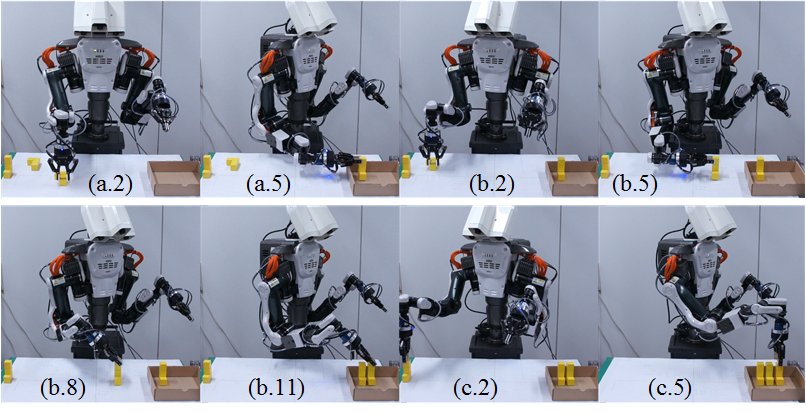}
	\caption{The real-world execution of packing objects. The identifiers (a.2),
	(a.5), \ldots are not continuous. They are used intentionally to help users
	find the correspondent simulation results in Fig.\ref{packingobjects}.}
	\label{packingobjectsreal}
\end{figure}

The planned results are sent to the Kawada Nextage robot for execution.
Fig.\ref{packingobjectsreal} shows the results of execution. The identifiers
(a.2), (a.5), \ldots are not continuous. They are used to help users
find the correspondent robot poses and grasp configurations in
Fig.\ref{packingobjects}.
	
\subsection{Assembly}

\begin{figure}[!htbp]
	\centering
	\includegraphics[width=.46\textwidth]{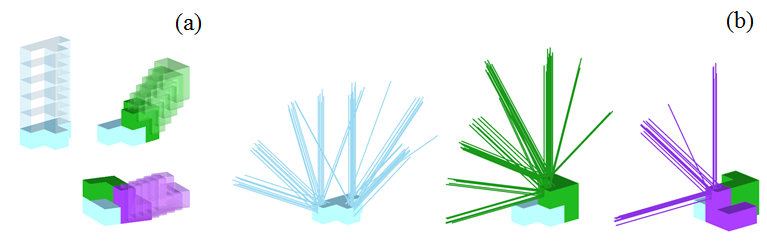}
	\caption{(a) The assembly sequence (assembly order and assembly directions)
	planned by a high level planning component. (b) The accessible grasps of each
	object following the assembly sequence in (a). The accessible grasps are
	computed by the grasp planner of the mid-level planning system.}
	\label{graspexpsim2}
\end{figure}

The mesh models used in the assembly task are shown in
Fig.\ref{graspexpsim2}. The goal is to assembly three blocks into a structure
shown in the lower part of Fig.\ref{graspexpsim2}(a). A high level planning
component computes an assembly sequence which will be sent to the mid-level
planning system as input.
The assembly sequence includes the assembly order, assembly directions, and goal
poses of the objects (Fig.\ref{graspexpsim2}(a)). The mid-level planner uses the
assembly sequence to compute the manipulation sequence of each object.

The first step is to compute the accessible grasps.
Fig.\ref{graspexpsim2}(b) shows
the accessible grasps of each object following the assembly sequence in
Fig.\ref{graspexpsim2}(a). The accessible grasps are computed by the grasp
planner of the mid-level planning system. Since the assembled parts become
obstacles of the remaining parts, the number of accessible grasps decrease as
assembly progresses.

\begin{figure}[!htbp]
	\centering
	\includegraphics[width=.46\textwidth]{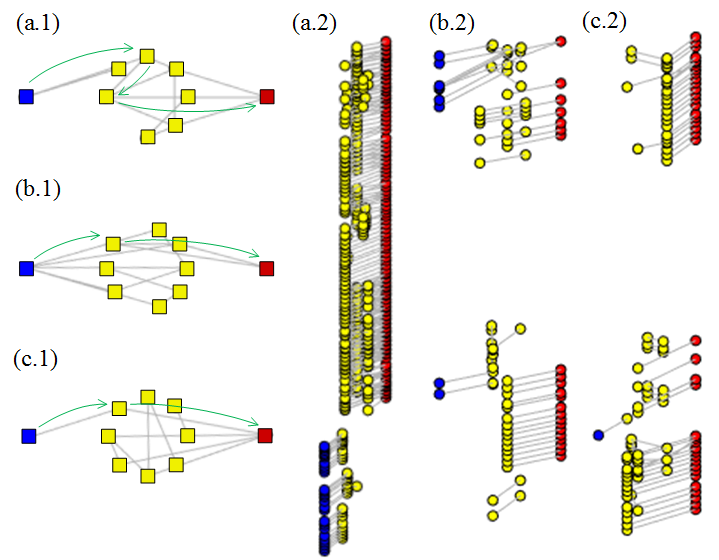}
	\caption{Regrasp graph of the three objects in the assembly task. (a.1) and
	(a.2) are the first layer and second layer of the regrasp graph for the
	first object. (b.1) and (b.2) are the first layer and second layer of the
	regrasp graph for the second object. (c.1) and (c.2) are the first layer and
	second layer of the regrasp graph for the third object.}
	\label{assemblyinggraphs}
\end{figure}

\begin{figure*}[!htbp]
	\centering
	\includegraphics[width=.96\textwidth]{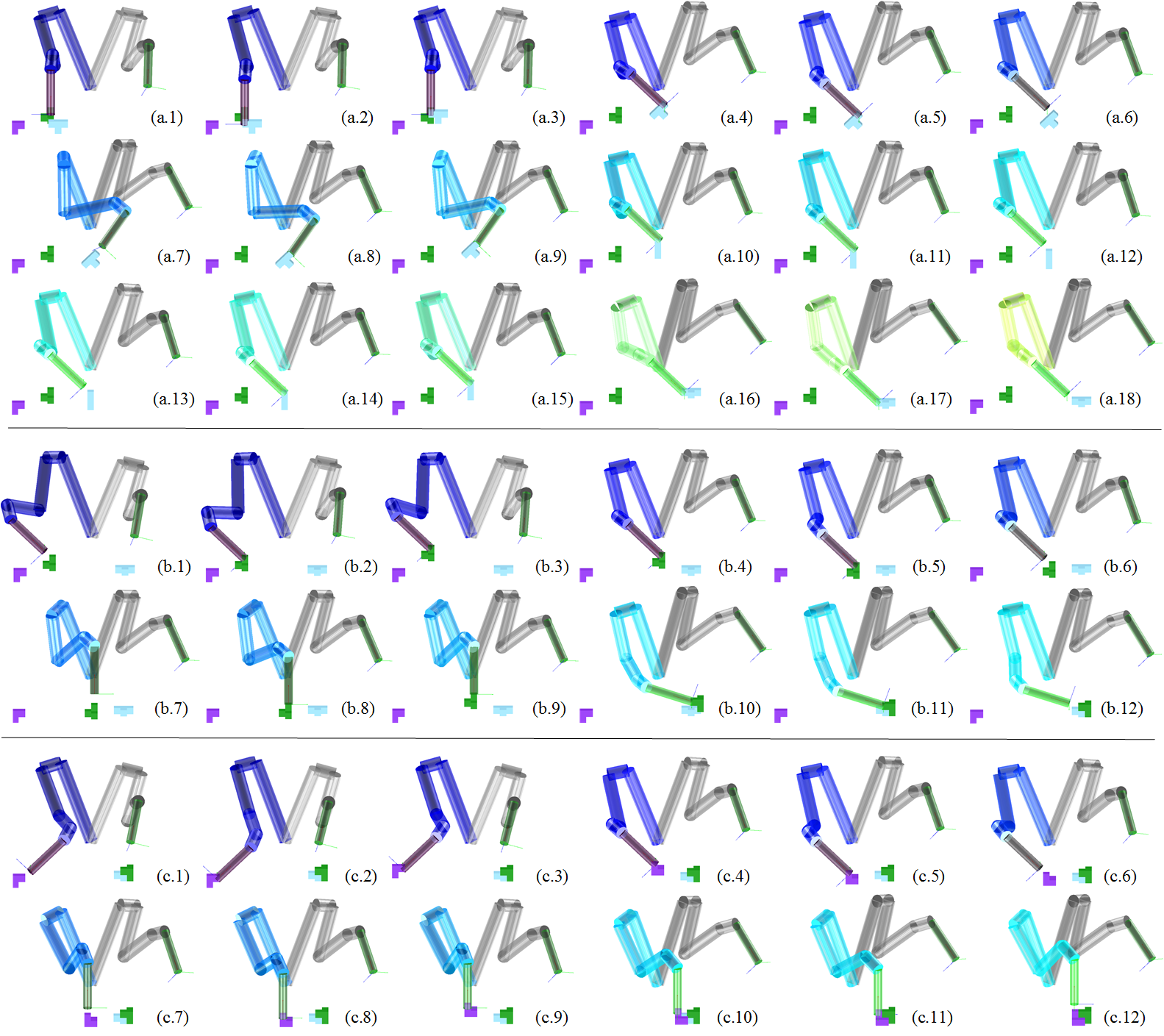}
	\caption{The simulation results of assemblying objects. (a.1), (a.2), \ldots,
	(a.18) are the robot poses and grasp configurations to assembly the first
	object. (b.1), (b.2), \ldots, (b.12) are the robot poses and grasp
	configurations to assembly the second object. (c.1), (c.2), \ldots,
	(c.12) are for assemblying the third object.}
	\label{assemblyingobjects}
\end{figure*}

Then, the placement planner computes the stable placements of each object and
the regrasp solver uses the stable placements and their accessible grasps to build
regrasp graphs. Fig.\ref{assemblyinggraphs} shows the regrasp graphs. Like
Fig.\ref{packingobjgraphs}, (a.1) and (a.2) are the first layer and second layer
of the regrasp graph for the first object. (b.1) and (b.2) are the first layer and second layer of the
regrasp graph for the second object. (c.1) and (c.2) are the first layer and
second layer of the regrasp graph for the third object. The green arrows show
the paths obtained by searching the first layer. The first object requires two
times of regrasp. The second and third objects both require one time of regrasp.
Fig.\ref{assemblyingobjects} shows the robot poses and grasp configurations
planned by the mid-level planning system. The green arrows in Fig.\ref{assemblyinggraphs}(a.1) correspond to
Fig.\ref{assemblyingobjects}(a.2)
$\rightarrow$ (a.5)(a.8) $\rightarrow$ (a.11)(a.14) $\rightarrow$ (a.17).
The green arrows in Fig.\ref{assemblyinggraphs}(b.1) correspond to
Fig.\ref{assemblyingobjects}(b.2) $\rightarrow$ (b.5)(b.8) $\rightarrow$ (b.11).
The green arrows in Fig.\ref{assemblyinggraphs}(c.1) correspond to
Fig.\ref{assemblyingobjects}(c.2) $\rightarrow$ (c.5)(c.8) $\rightarrow$ (c.11).
The remaining robot poses and grasp configurations are the motion primitives.

Note that in assembly planning, the $backward$ directions of
placing-down motion primitives are decided by the assembly directions planned by
the high level component. This is different from packing where $backward$ is
defined as the $upward$ direction. The Tool Point Center (TCP) of the active arm
in the motion primitives of
Fig.\ref{packingobjects}(a.4) $\rightarrow$ (a.5), (a.10) $\rightarrow$ (a.11),
(b.4) $\rightarrow$ (b.5), and (c.4) $\rightarrow$ (c.5) moves downwards. In
contrast, the TCP of the active arm in the motion primitives
Fig.\ref{assemblyingobjects}(a.16) $\rightarrow$ (a.17),
(b.10) $\rightarrow$ (b.11), and (c.10) $\rightarrow$ (c.11) moves along the
planned directions shown in Fig.\ref{graspexpsim2}(a).

\begin{figure*}[!htbp]
	\centering
	\includegraphics[width=.98\textwidth]{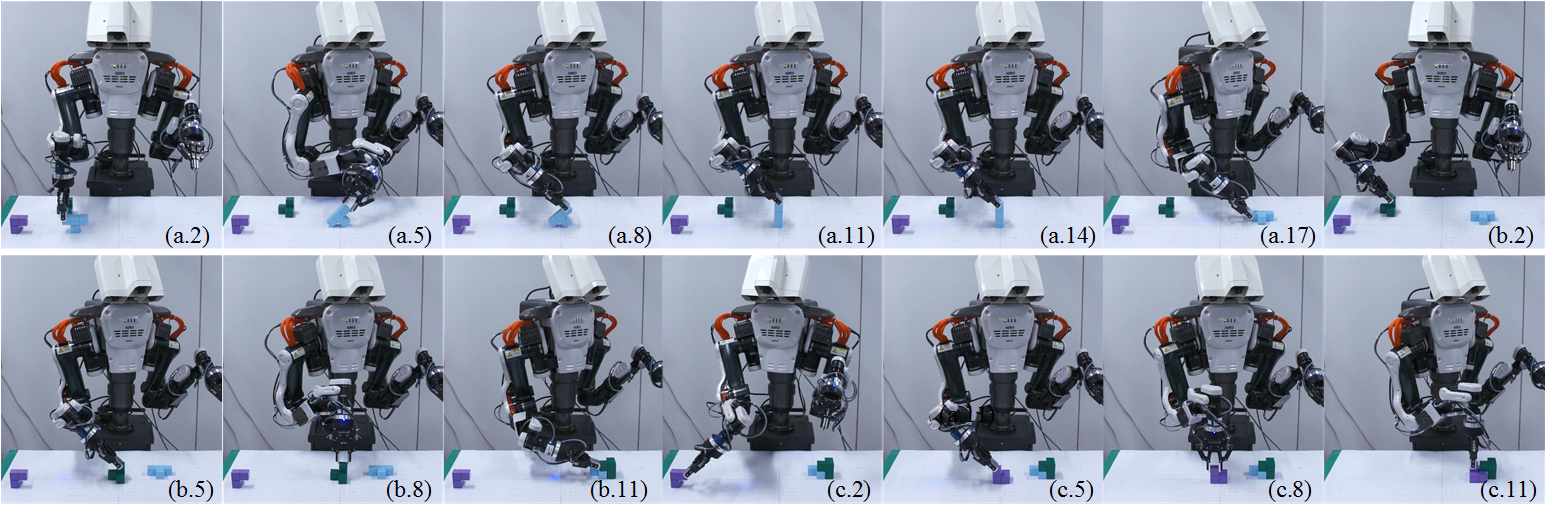}
	\caption{The real-world execution of assembly. The identifiers (a.2),
	(a.5), \ldots corresponds to the counterpart in	Fig.\ref{assemblyingobjects}.}
	\label{assemblyingobjectsreal}
\end{figure*}

Like the packing task, the planned results are sent to the Kawada Nextage robot
for execution. Fig.\ref{assemblyingobjectsreal} shows the results of execution. The
identifiers (a.2), (a.5), \ldots relate the real-world snapshots to robot poses
and grasp configurations in Fig.\ref{packingobjects}.

\subsection{Using tools}

\begin{figure*}[!htbp]
	\centering
	\includegraphics[width=.98\textwidth]{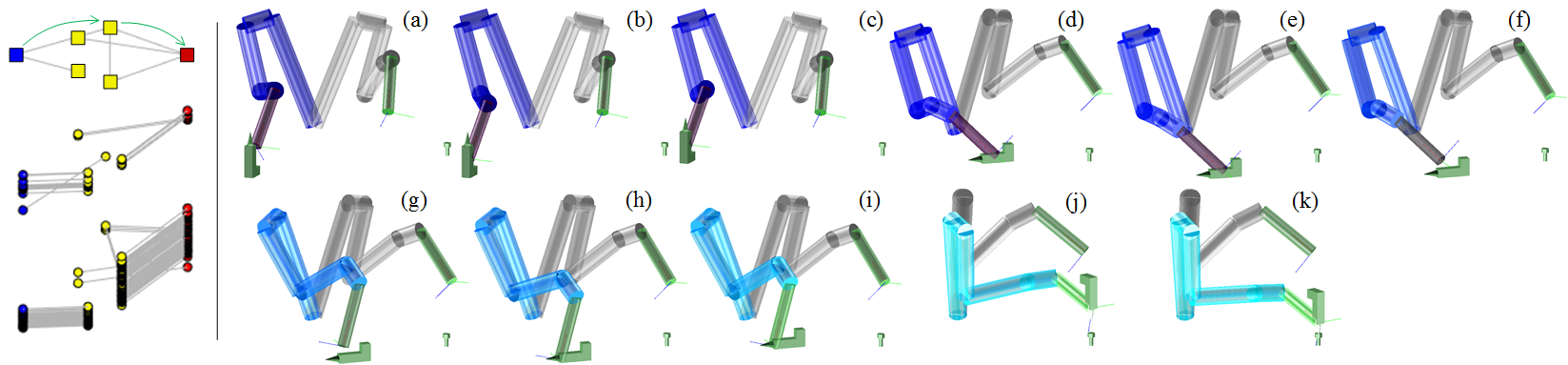}
	\caption{Regrasp graph of drill and the simulation result. The upper-left
	graph with rectangular nodes is the first layer of the regrasp graph. The
	lower-left graph with circular nodes is the second layer of the regrasp graph.
	The remaining (a), (b), \ldots, (k) show the sequence of robot poses and grasp
	configurations computed by the mid-level planning system.}
	\label{usingtools}
\end{figure*}

The tool used in this experiment is a drill and the robot is expected to move
the drill to fasten a nut. A high-level planner specifies the goal pose of the
drill and the mid-level planning system computes a sequence of robot poses
and grasp configurations that move the drill from its initial pose to the goal.
Fig.\ref{usingtools} shows the regrasp graph of the drill and the simulation
result. The initial and goal poses of the drill are shown in
Fig.\ref{usingtools}(a) and (k).
The output of the graph search in the first layer is
(b) $\rightarrow$ (c) $\rightarrow$ (h) $\rightarrow$ (k). After expansion, the
output is (a) $\rightarrow$ (b) $\rightarrow$ \ldots $\rightarrow$ (k) as shown
in the figure.

\begin{figure}[!htbp]
	\centering
	\includegraphics[width=.48\textwidth]{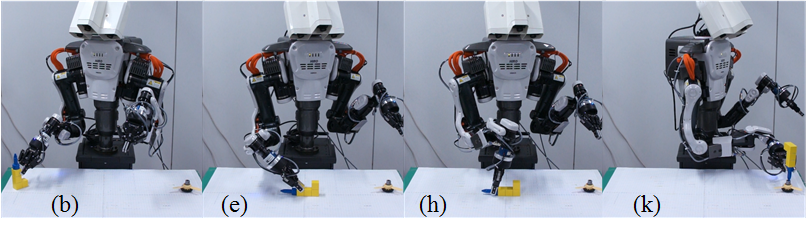}
	\caption{The real-world execution of using a tool. The identifiers (b),
	(e), \ldots corresponds to the counterpart in Fig.\ref{usingtools}.}
	\label{usingtoolsreal}
\end{figure}

Fig.\ref{usingtoolsreal} shows the results of real-world execution. The
identifiers (a.2), (a.5), \ldots relate the real-world snapshots to robot poses
and grasp configurations in Fig.\ref{packingobjects}.
	
\subsection{Cost analysis}

The time cost of the three experiments are shown in Table.I.
The values are obtained by running the algorithms on the Thinkpad P70 mobile
workstation using Matlab 2016a. The operating system
is Windows 10.

The columns named
``Grasps'', ``Placements'', and ``Regrasp grasph solver'' show the cost of the
three planners in the mid-level planning system, respectively. 
The cells filled with ``-'' indicate their values are the same as the cells on
top of them. The ``IK,CD(init)'', ``IK,CD(regrasp)'', and
``IK,CD(goal)'' under the ``Regrasp grasph solver'' column show the cost to
compute the feasibility of IK and check the collision between the robot and obstacles at the initial object
configurations, the goal object configurations, and the regrasp positions,
respectively. Both original pose and the poses at the motion primitives are
examined. The ``IK,CD(total)'' column is the sum of the values in
``IK,CD(init)'', ``IK,CD(regrasp)'', and ``IK,CD(goal)''. The ``Graph
search'' column shows the cost to build the two layer regrasp graph, search the
graph, and generate a sequence of robot poses and grasp configurations by
expanding motion primitives.
It is essentially the cost of the algorithm shown in Alg.\ref{searchexp}.

The overall cost of all planners is around 300$s$, which implies that the
algorithms cannot be used online. However, some of the data could be
pre-computed offline and reused during execution. These data are marked with
gray color. The grasps, the placements, and the IK and CD at the regrasp
positions and goal object configurations are usually pre-defined. They can be
pre-computed offline.
The IK and CD at the initial object configurations must be done online since
the initial object configurations are obtained from vision systems and change
from time to time. The distribution of on-line and off-line computation load
depends on users. Higher on-line load makes the system more flexible. For
instance, making the IK and CD at the regrasp configuration online allows users
to change the positions to do regrasp during execution. However, the on-line
cost increases. Lower on-line load, in the extreme case, reduces time cost to
less than 30$s$ (the ``IK,CD(init)'' column plus the ``Graph search'' column)
which can be used online.

\begin{table*}
\centering
\renewcommand{\arraystretch}{1.2}
\caption{\label{conf0}Time efficiency of the three experiments}
%\resizebox{0.99\linewidth}{!}{%
\begin{tabular}{ccccccccc}
\toprule
 & & \multirow{ 2}{*}{{\color{chgray}Grasps}} & \multirow{
 2}{*}{{\color{chgray}Placements}} &
 \multicolumn{5}{c}{{\color{chgray}Regrasp graph solver}}
 \\
 & & &  & IK,CD(init) & {\color{chgray}IK,CD(regrasp)} & 
 {\color{chgray}IK,CD(goal)} &
 {\color{chgray}IK,CD(total)} & Graph search\\
 \midrule
 \midrule
 \multirow{ 3}{*}{Packing} & 1st object & {\color{chgray}16.865205$s$} &
 {\color{chgray}108.958532$s$} & 8.671939$s$ & {\color{chgray}100.330880$s$} &
{\color{chgray} 31.503069$s$} & {\color{chgray}142.877699$s$} & 2.371811$s$\\
 & 2nd object & {\color{chgray}-} & {\color{chgray}-} & 9.024444$s$ &
 {\color{chgray}-} & {\color{chgray}9.836328$s$} & {\color{chgray}116.819841$s$}
 & 0.547989$s$\\
 & 3rd object & {\color{chgray}-} & {\color{chgray}-} & 40.426119$s$ &
 {\color{chgray}-} & {\color{chgray}9.733597$s$} & {\color{chgray}150.490596$s$}
 & 0.606657$s$
 \\
 \multirow{ 3}{*}{Assembly} & 1st object & {\color{chgray}16.139938$s$} &
 {\color{chgray}159.130246$s$} & 11.199010$s$ & {\color{chgray}171.522333$s$} &
 {\color{chgray}5.736903$s$} & {\color{chgray}189.249336$s$} & 0.874507$s$ \\
 & 2nd object & {\color{chgray}17.385174$s$} & {\color{chgray}104.939474$s$} &
 8.990426$s$ & {\color{chgray}70.594335$s$} & {\color{chgray}3.016662$s$} &
 {\color{chgray}82.601423$s$} & 0.496157$s$ \\
 & 3rd object & {\color{chgray}16.984532$s$} & {\color{chgray}104.457831$s$} &
 18.872237$s$ & {\color{chgray}80.420244$s$} & {\color{chgray}1.561027$s$} &
 {\color{chgray}100.853488$s$} & 0.448476$s$ \\
 Using a tool & tool object & {\color{chgray}33.855902$s$} &
 {\color{chgray}131.262263$s$} & 62.269052$s$ & {\color{chgray}108.614137$s$} &
 {\color{chgray}93.858813$s$} & {\color{chgray}264.742002$s$} & 0.492925$s$ \\
\midrule
\bottomrule
\end{tabular}
%}
\end{table*}

\section{Conclusions and Future Work}

A mid-level planning system is presented in this paper. The input to the system
is a sequence of object poses planned by a high-level assembly or symbolic
planning component. The output of the system is a sequence of robot poses and
grasp configurations that will be used by a low-level motion planning component.
The planning system is composed of a grasp planner, a placements planner, and a regrasp sequence
solver. The paper presented the details of these planners and solvers, and
demonstrated their efficacy using three exemplary tasks. 

The current implementation of the demonstrations orchestrate the planned
robot poses and grasps using hard-coded interpolations. The applications are
limited to small-scale wooden blocks. In the future, the mid-level planning
system with be integrated with high-level and low-level components to challenge
more pragmatic tasks.

\section*{Acknowledgment}
This paper is based on results obtained from a project commissioned by the
New Energy and Industrial Technology Development Organization (NEDO).

\ifCLASSOPTIONcaptionsoff
  \newpage
\fi



\bibliographystyle{IEEEtran}
\bibliography{reference}

\begin{thebibliography}{10}
\providecommand{\url}[1]{#1}
\csname url@samestyle\endcsname
\providecommand{\newblock}{\relax}
\providecommand{\bibinfo}[2]{#2}
\providecommand{\BIBentrySTDinterwordspacing}{\spaceskip=0pt\relax}
\providecommand{\BIBentryALTinterwordstretchfactor}{4}
\providecommand{\BIBentryALTinterwordspacing}{\spaceskip=\fontdimen2\font plus
\BIBentryALTinterwordstretchfactor\fontdimen3\font minus
  \fontdimen4\font\relax}
\providecommand{\BIBforeignlanguage}[2]{{%
\expandafter\ifx\csname l@#1\endcsname\relax
\typeout{** WARNING: IEEEtran.bst: No hyphenation pattern has been}%
\typeout{** loaded for the language `#1'. Using the pattern for}%
\typeout{** the default language instead.}%
\else
\language=\csname l@#1\endcsname
\fi
#2}}
\providecommand{\BIBdecl}{\relax}
\BIBdecl

\bibitem{Mello:1990wqba}
L.~H. de~Mello, ``{AND/OR Graph Representation of Assembly Plans},'' \emph{IEEE
  Transaction on Robotics and Automation}, 1990.

\bibitem{Mello91}
A.~C.~S. LS~Hommem De~Mello, ``{A Correct and Complete Algorithm for the
  Generation of Mechanical Assembly Sequences},'' \emph{IEEE Transactions on
  Robotics and Automation}, 1991.

\bibitem{Wilson94}
R.~Wilson and J.-C. Latombe, ``{Geometric Reasoning About Mechanical
  Assembly},'' \emph{{Artificial Intelligence}}, 1994.

\bibitem{McDermott98}
M.~Ghallab \emph{et~al.}, ``{PDDL -- The Planning Domain Definition Language --
  Version 1.2},'' Yale Center for Computational Vision and Control, Tech. Rep.,
  1998.

\bibitem{Hoffmann06}
J.~Hoffmann, S.~Edelkamp, S.~Thiebaux, R.~Englert, F.~dos Santos~Liporace, and
  S.~Trug, ``{Engineering Benchmarks for Planning: the Domains Used in the
  Deterministic Part of IPC-4},'' \emph{Journal of Artificial Intelligence
  Research}, 2006.

\bibitem{Kavraki96}
L.~Kavraki, P.~Svestka, J.~C. Latombe, and M.~Overmars, ``{Probabilistic
  Roadmaps for Path Planning in High-Dimensional Configuration Spaces},''
  \emph{IEEE Transactions on Robotics and Automation}, vol.~12, pp. 566--580,
  1996.

\bibitem{Lavalle00}
S.~M. Lavalle and J.~J. Kuffner, ``{Rapidly-Exploring Random Trees: Progress
  and Prospects},'' in \emph{Proceedings of International Workshop on the
  Algorithmic Foundations of Robotics}, 2000, pp. 293--308.

\bibitem{Zucker12}
M.~Zucker, N.~Ratliff, A.~D. Dragan, M.~Pivtoraiko, M.~Klingensmith, C.~M.
  Dellin, J.~A. Bagnell, and S.~S. Srinivasa, ``{CHOMP: Covariant Hamiltonian
  Optimization for Motion Planning},'' \emph{International Journal of Robotic
  Research (IJRR)}, 2012.

\bibitem{Schulman14}
J.~Schulman, Y.~Duan, J.~Ho, A.~Lee, I.~Awwal, H.~Bradlow, J.~Pan, S.~Patil,
  K.~Goldberg, and P.~Abbeel, ``{Motion Planning with Sequential Convex
  Optimization and Convex Collision Checking},'' \emph{International Journal of
  Robotic Research (IJRR)}, 2014.

\bibitem{Fikes71}
R.~E. Fikes and N.~J. Nilsson, ``{STRIPS: A New Approach to the Application of
  Theorem Proving to Problem Solving},'' \emph{Artificial Intelligence}, 1971.

\bibitem{Beta07}
C.~Belta, A.~Bicchi, M.~Egerstedt, E.~Frazzoli, E.~Klavins, and G.~J. Pappas,
  ``{Symbolic Plannign and Control of Robot Motion},'' \emph{IEEE Robotics and
  Automation Magzine}, 2007.

\bibitem{Lahijanian16}
M.~Lahijanian, M.~R. Maly, D.~Fried, L.~E. Kavraki, H.~Kress-Gazit, and M.~Y.
  Vardi, ``{Iterative Temporal Planning in Uncertain Environments with Partial
  Satisfaction Guarantees},'' \emph{Transaction on Robotics}, 2016.

\bibitem{Guo14}
M.~Guo and D.~V. Dimarogonas, ``{Multi-agent Plan Reconfiguration under Local
  LTL Specifications},'' \emph{International Journal of Robotics Research},
  2014.

\bibitem{KG09}
H.~Kress-Gazit, G.~E. Fainekos, and G.~J. Pappas, ``{Temporal Logic-based
  Reactive Mission and Motion Planning},'' \emph{IEEE Transaction on Robotics},
  2009.

\bibitem{Hertle11}
A.~Hertle, ``{Design and Implementation of an Object-Oriented Planning
  Lanuage},'' Master's thesis, Albert-Ludwigs-Universitat reiburg, 2011.

\bibitem{Dogar15}
M.~Dogar, A.~Spielberg, S.~Baker, and D.~Rus, ``{Multi-Robot Grasp Planning for
  Sequential Assembly Operations},'' in \emph{Proceedings of IEEE International
  Conferene on Robotics and Automation (ICRA)}, 2015.

\bibitem{Bidot2015}
J.~Bidot, L.~Karlsson, F.~Lagriffoul, and A.~Saffiotti, ``{Geometric
  Backtracking for Combined Task and Motion Planning in Robotic systems},''
  \emph{Artificial Intelligence}, 2015.

\bibitem{Knepper:2013fn}
R.~Knepper, T.~Layton, J.~Romanishin, and D.~Rus, ``{IkeaBot: An Autonomous
  Multi-robot Coordinated Furniture Assembly System},'' in \emph{Proceedings of
  IEEE International Conference on Robotics and Automation (ICRA)}, 2013.

\bibitem{Stein11}
D.~Stein, T.~R. Schoen, and D.~Rus, ``{Constraint-aware Coordinated
  Construction of Generic Structures},'' in \emph{Proceedings of IEEE
  International Conference on Intelligent Robots and Systems}, 2011.

\bibitem{Knepper:2012}
R.~Knepper and D.~Rus, ``{Pedestrian-inspired Sampling-based Multi-robot
  Collision Avoidance},'' in \emph{Proceedings of IEEE International Symposium
  on Robot and Human Interactive Communication}, 2012.

\bibitem{Kaelbling13}
L.~P. Kaelbling and T.~Lozano-Perez, ``{Integrated Task and Motion Planning in
  Belief Sapce},'' \emph{International Journal of Robotics Research}, 2013.

\bibitem{Dantam13}
N.~T. Dantam, ``{The Motion Grammar: Analysis of a Linguistic Method for Robot
  Control},'' \emph{IEEE Transactions on Robotics}, 2013.

\bibitem{Kris10}
K.~Hauser and J.-C. Latombe, ``{Multi-modal Motion Planning in Non-expansive
  Spaces},'' \emph{International Journal of Robotics Research}, pp. 897--915,
  2010.

\bibitem{Dantam16}
N.~Dantam, Z.~Kingston, and S.~Chauhuri, ``{Incremental Task and Motion
  Planning: A Constraint-Based Approach},'' in \emph{Proceedings of Robotics:
  Science and Systems (RSS)}, 2016.

\bibitem{Bekris16}
A.~Krontiris and K.~Bekris, ``{Efficiently Solving General Rearrangement Tasks:
  A Fast Extension Primitive For An Incremental Sampling Based Planner},'' in
  \emph{Proceedings of IEEE International Conference on Robotics and Automation
  (ICRA)}, 2016.

\bibitem{ICRA2014:Nedunuri}
S.~Nedunuri, S.~Prabhu, and M.~Moll, ``{SMT-based Synthesis of Integrated Task
  and Motion Plans from Plan Outlines},'' in \emph{{Proceedings of IEEE
  International Conference on Robotics and Automation (ICRA)}}, 2014.

\bibitem{King2013}
J.~King, M.~Klingensmith, and C.~Dellin, ``{Regrasp Manipulation as Trajectory
  Optimization},'' in \emph{Proceedings of Robotics: Science and Systems
  (RSS)}, 2013.

\bibitem{Heger10}
F.~W. Heger, ``{Assembly Planning in Constrained Environments: Building
  Structures with Multiple Mobile Robots},'' Ph.D. dissertation, Carnegie
  Mellon University, 2010.

\bibitem{Eiichi10}
E.~Yoshida, C.~Esteves, O.~Kanoun, M.~Poirier, A.~Mallet, J.-P. Laumond, and
  K.~Yokoi, ``{Planning Whole-body Humanoid Locomotion, Reaching and
  Manipulation},'' \emph{Motion Planning for Humanoid Robots}, pp. 99--128,
  2010.

\bibitem{Pierre87}
P.~Tournassound, T.~Lozano-Perez, and E.~Mazer, ``{Regrasping},'' in
  \emph{Proceedings of International Conference on Robotics and Automation},
  1987, pp. 1924--1928.

\bibitem{Rohrdanz97}
F.~Rohrdanz and F.~M. Wahl, ``{Generating and Evaluating Regrasp Operations},''
  in \emph{Proceedings of International Conference on Robotics and Automation},
  1997, pp. 2013--2018.

\bibitem{Sascha99}
S.~A. Stoeter, S.~Voss, N.~P. Papanikolopoulos, and H.~Mosemann, ``{Planning of
  Regrasp Operations},'' in \emph{Proc. of ICRA}, 1999, pp. 245--250.

\bibitem{Hajime98}
H.~Terasaki and T.~Hasegawa, ``{Motion Planning of Intelligent Manipulation by
  a Parallel Two-Fingered Gripper Equiped with a Simple Rotating Mechanism},''
  \emph{IEEE Transaction on Robotics and Automation}, 1998.

\bibitem{Cho03}
K.~Cho, M.~Kim, and J.-B. Song, ``{Complete and Rapid Regrasp Planning with
  Look-up Table},'' \emph{Journal of Intelligent and Robotic Systems}, pp.
  371--387, 2003.

\bibitem{Repela02}
D.~R. Rapela and U.~Rembold, ``{Planning of Regrasping Operations for a
  Dexterous Hand in Assembly Tasks},'' \emph{Journal of Intelligent and Robotic
  Systems}, vol.~33, pp. 231--266, 2002.

\bibitem{Zhixing08}
Z.~Xue \emph{et~al.}, ``{Planning Regrasp Operations For A Multifingered
  Robotic Hand},'' in \emph{Proceedings of International Conference on
  Automation Sicence and Engineering}, 2008, pp. 778--783.

\bibitem{Lert16}
P.~Lertkultanon and Q.-C. Pham, ``{A single-query manipulation planner},''
  2016.

\bibitem{Yoshihito92}
Y.~Koga and J.-C. Latombe, ``{Experiments in Dual-arm Manipulation Planning},''
  in \emph{Proceedings of IEEE International Conference on Robotics and
  Automation (ICRA)}, 1992, pp. 2238--2245.

\bibitem{Yoshihito94a}
------, ``{On Multi-Arm Manipulation Planning},'' in \emph{Proceedings of IEEE
  International Conference on Robotics and Automation (ICRA)}, 1994, pp.
  945--952.

\bibitem{Niko09}
N.~Vahrenkamp, D.~Berenson, T.~Asfour, J.~Kuffner, and R.~Dillmann, ``{Humanoid
  Motion Planning for Dual-arm Manipulation and Regrasp Tasks},'' in
  \emph{Proceedings of IEEE/RSJ International Conference on Intelligent Robots
  and Systems (IROS)}, 2009.

\bibitem{Benjamin10}
B.~Cohen \emph{et~al.}, ``{Planning Single-arm Manipulations with N-Arm
  Robots},'' in \emph{Proceedings of Robotics: Science and Systems}, 2010.

\bibitem{Harada14}
K.~Harada \emph{et~al.}, ``Project on development of a robot system for random
  picking-grasp/manipulation planner for a dual-arm manipulator,'' in
  \emph{IEEE/SICE International Symposium on System Integration (SII)}, 2014,
  pp. 583--589.

\bibitem{Wan2015a}
W.~Wan, M.~T. Mason, R.~Fukui, and Y.~Kuniyoshi, ``Improving regrasp algorithms
  to analyze the utility of work surfaces in a workcell,'' in \emph{Proceedings
  of International Conference on Robotics and Automation}, 2015.

\bibitem{Wan2015c}
W.~Wan and K.~Harada, ``{Reorientating Objects with a Gripping Hand and a Table
  Surface},'' in \emph{Proceedings of international Conference on Humanoid
  Robots}, 2015.

\bibitem{Chao16}
C.~Cao, W.~Wan, J.~Pan, and K.~Harada, ``{Analyzing the Utility of a Support
  Pin in Sequential Robotic Manipulation},'' in \emph{Proceedings of IEEE
  International Conference on Robotics and Automation (ICRA)}, 2016.

\bibitem{Jean10}
J.-P. Saut, M.~Gharbi, J.~Cortes, D.~Sidobre, and T.~Simeon, ``{Planning
  Pick-and-place Tasks with Two-hand Regrasping},'' in \emph{Proceedings of
  IEEE/RSJ International Conference on Intelligent Robots and Sytems (IROS)},
  2010.

\bibitem{Wan2016ral}
W.~Wan and K.~Harada, ``{Developing and Comparing Single-arm and Dual-arm
  Regrasp},'' \emph{IEEE Robotics and Automation Letters}, 2016.

\bibitem{Wan2016ar}
------, ``{Achieving High Success Rate in Dual-arm Handover Using Large Number
  of Candidate Grasps},'' \emph{Advanced Robotics}, 2016.

\bibitem{Hauser06}
K.~K. Hauser, T.~Bretl, K.~Harada, and J.-C. Latombe, ``{Using Motion
  Primitives in Probabilistic Sample-Based Planning for Humanoid Robots},'' in
  \emph{Proceedings of Robotics: Science and Systems (RSS)}, 2006.

\bibitem{Ding13}
X.~Ding and C.~Fang, ``{A Novel Method of Motion Plnning for an Anthropomorphic
  Arm Based on Movement Primitives},'' \emph{IEEE Transactions on
  Mechatronics}, 2013.

\bibitem{Hsu03}
D.~Hsu, T.~Jiang, J.~Reif, and Z.~Sun, ``{The Bridge Test for Sampling Narrow
  Passages with Probabilistic Roadmap Planners},'' in \emph{Proceedings of IEEE
  International Conference on Robotics and Automation (ICRA)}, 2003.

\bibitem{Yershova2009}
A.~Yershova and S.~M. LaValle, ``{Motion Planning for Highly Constrained
  Spaces},'' in \emph{Robot Motion and Control 2009}.\hskip 1em plus 0.5em
  minus 0.4em\relax Springer, 2009, pp. 297--306.

\end{thebibliography}
\end{document}